\documentclass{article}
\usepackage{spconf,amsmath,graphicx,amssymb}
\usepackage{hyperref}
\usepackage{booktabs}
\usepackage{amsfonts}
\usepackage{color}
\usepackage{multirow}
\usepackage{enumitem}
\usepackage[table,xcdraw]{xcolor}
\usepackage{float}
\usepackage{bm}
\usepackage[normalem]{ulem}
\usepackage[ruled]{algorithm2e}
\usepackage{caption}
\usepackage{subcaption}
\usepackage[symbol, multiple]{footmisc}
\graphicspath{{images/}}

\title{SP-BatikGAN: An Efficient Generative Adversarial Network for Symmetric Pattern Generation}
\name{$^{\star \dagger \ddagger}$Chrystian\thanks{During this research the author is a receipient to Sea Scholarship Indonesia and the Hanrieder Foundation for Excellence by Max Planck Foundation.}, $^{\ddagger}$Wahyono}
\address{
$^{\star}$Samsung Research Indonesia, Jakarta\\
$^{\dagger}$Max Planck Institute for Biological Cybernetics, Max Plank Society, Tübingen\\
$^{\ddagger}$Department of Computer Science and Electronics, Gadjah Mada University, Yogyakarta}

\begin{document}
%
\maketitle
\begin{abstract}



Following the contention of AI arts, our research focuses on bringing AI for all, particularly for artists, to create AI arts with limited data and settings.
We are interested in geometrically symmetric pattern generation, which appears on many artworks such as Portuguese, Moroccan tiles, and Batik, a cultural heritage in Southeast Asia. 
Symmetric pattern generation is a complex problem, with prior research creating too-specific models for certain patterns only. 
We provide publicly\footnote[1]{ITB-mBatik Dataset with CC-BY-NC-SA license can be found:\\ \url{https://data.mendeley.com/datasets/7hzr5539ws}}, the first-ever 1,216 high-quality symmetric patterns straight from design files for this task.
We then formulate symmetric pattern enforcement (SPE) loss to leverage underlying symmetric-based structures that exist on current image distributions. 
Our SPE improves and accelerates training on any GAN configuration, and, with efficient attention, SP-BatikGAN compared to FastGAN, the state-of-the-art GAN for limited setting, improves the FID score from \textbf{110.11} to \textbf{90.76}, an \textbf{18\%} decrease, and model diversity recall score from \textbf{0.047} to \textbf{0.204}, a \textbf{334\%} increase.

\end{abstract}
\begin{keywords}
GAN, Application, Unsupervised Image Generation, Batik Generation, Symmetric Pattern Generation
\end{keywords}
\section{Introduction}
\label{sec:intro}


Batik is a cultural icon in Southeast Asia, a `Masterpiece of Oral and Intangible Heritage of Humanity' by UNESCO, is an ancient art to produce beautiful textiles. 
In recent year many research \cite{9527514, 9527490, 10.1145/3379173.3393710, 8981834, Esyenne2019, 8978233} interests grew utilizing AI to generate art with unique complex pattern style such as batik.
However, this early preliminearies research suffer from difficulty of training unstable GAN, the need of many data sample, the difficulty to get high quality data suitable for unsupervised image generation, and pattern structure from each art is complex and vary greatly.

First problem, while large dataset is preferable and important, but also too the quality of the images. Second, even when we have the dataset, mixing different pattern type or style together does not necessarily give a guarentee of good results i.e. even with thousands of images, result still not as high quality as it could be\cite{9527514, 8981834}. 
Recent research \cite{10.1145/3379173.3393710} tradeoff generality with quality as their approach is to limit a given style with a fixed loss function only for a single specific pattern type.

To address these problems, we first collect and preprocess high-quality batik data. Then, we focused on generalizing symmetrical pattern generation tasks. We outline a framework that allows the generative model to efficiently train the generator on symmetric pattern tasks, which is one of the key characteristics of batik and other cultural artwork as well. From the framework itself, we then intuitively derive \textbf{Symmetric Pattern Enforcement (SPE)} Loss. With SPE we are able to improve image quality and detail from state-of-the-art FastGAN\cite{liu2021faster} on symmetric pattern generation task.

In short, our contributions are the following:
\begin{itemize}
	\item We generalize symmetric pattern generation task for other different symmetric pattern type (Section \ref{section:spg})
	\item We create intuitive learning method by deriving SPE loss from existing symmetric structure in the dataset (Section \ref{section:spe})
	\item SPE loss improve the SOTA FastGAN and combined with Efficient Attention, SPE Loss accelerate GAN training, reducing FID, and drastically improve recall (Section \ref{section:accelerating})
\end{itemize}

\section{RELATED WORK}



\textbf{Prior related batik generation research}. In recent years there are great interest in batik generation with AI topics, such as \cite{8978233}, \cite{Esyenne2019}, and \cite{9527490} where they respectively use text-to-image, Conditional GAN, and neural style transfer. For unsupervised generation tasks, early research such as \cite{8981834}, uses simple DC-GAN to generate batik motif synthesis. Other \cite{9527514} continue this research with a more diverse dataset. However, for both research, they never measure any metrics and the generated image only has vague batik characteristics with too much noise. The most recent research for unsupervised batik generation tasks was BatikGAN \cite{10.1145/3379173.3393710}, however, BatikGAN has too many constraints. Even with five new custom losses, it only takes into account one type of pattern. The generated image is also low in resolution only 32×32 pixels per patch.

\textbf{Advancement on GAN for limited computation and dataset}. Unsupervised image generation GAN, from the first original GAN to the current SOTA at the time of writing such as StyleGAN2 \cite{Karras2019stylegan2}, the amount of data needed to train a GAN model is enormous. However, recent advancements such as Adaptive Discriminator Augmentation (ADA) \cite{Karras2020ada} and Differentiable Augmentation for data-efficient GAN \cite{zhao2020differentiable} can reduce the requirement of GAN data into only hundreds of images. This leads us to use the state-of-the-art FastGAN \cite{liu2021faster} as our baseline, where it outperforms previous SOTA StyleGAN2 in limited computation and dataset environments.

\section{METHODS}
\begin{figure}
	\centering
	\includegraphics[width=\columnwidth]{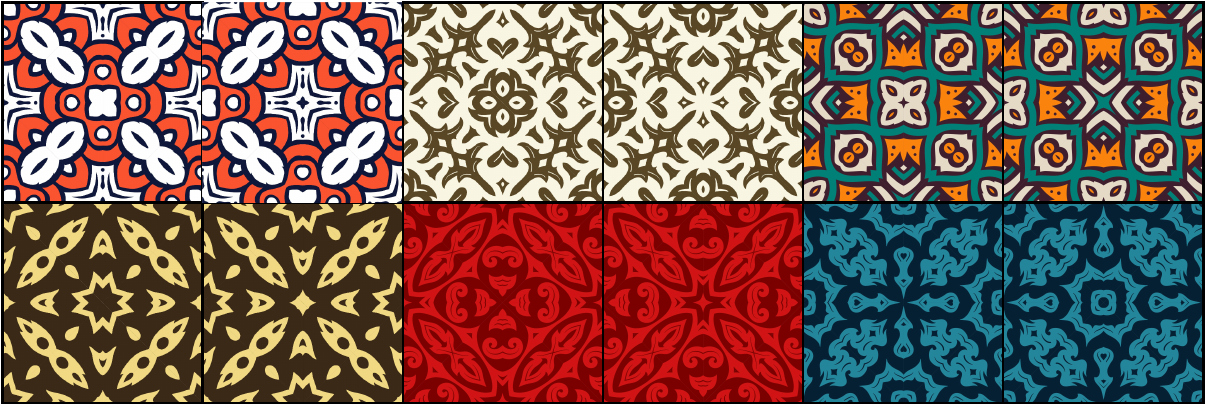}
	\caption[ITB-mBatik Dataset]{ITB-mBatik Dataset. Examples of the seamless and symmetric pattern pair, out of the final 1,216 in the final proposed dataset.}
	\label{fig:datadetail}
\end{figure}

To have a good image generation result, the quality of input images is crucial. 
In our case of Batik pattern, a traditional clothing pattern is difficult to acquire, especially in high resolution, and this is where most of the previous research was lacking. Previous research used UI Batik Dataset \cite{Gultom2018} and the image was taken only with a camera and was noticeably substandard.
Thus, for future research, we acquire and create our own more modern digital Batik dataset. We propose a better, new high-quality, publicly available dataset, straight from design files called \textit{Bandung Institute of Technology modern Batik} (ITB-mBatik) dataset \cite{https://doi.org/10.17632/7hzr5539ws.1}, a better-suited dataset for unsupervised image generation task shown in Fig. \ref{fig:datadetail}.

\subsection{Inefficiency of Whole Pattern Generation and Differing Symmetry Type}

Not only high-quality images are difficult to collect, another discernible problem in previous work \cite{9527514, 8981834} is the inefficiency of training the generator directly on a whole pattern. Fitting the generator with a full repeating pattern unnecessarily increase real image distribution complexity, where the generator needs to take to account many different arbitrary repeating pattern, resulting in poor subpar results. Let a "patch" be defined as the most minimum image subset, $\mathbf{p} \subset \mathbf{P}$ in which the patch can reconstruct the original repeating pattern, $\textbf{tile}(\mathbf{p}) = \mathbf{P}$. If a repeating pattern $\mathbf{P}$ can be fully reconstructed with the repetition of the subset $\mathbf{p}$, then rather than redundantly training the generator to learn the whole repeating pattern, generating the subset pattern is a better efficient approach. 

For patch generation, the patch patterns themselves have a high degree of freedom and flexibility. In addition to generating a patch, to limit and make the patch sensible to work with, we put a constraint that every patch across the dataset need to have a common symmetrical transformation. Training GAN with different symmetry will cause a problem of asymmetry and can damage the aesthetic of the output image. Therefore, putting a constraint on the dataset to have at least one similar symmetrical transformation preserve and avoid asymmetry.

\subsection{Symmetric Patch Pattern Generation Task within 2-Dimensional Basic Isometries}
\label{section:spg}

We define our tasks within the most common group of transformation, which is the Euclidian group within 2-D space or isometries of E(2) (keep in mind there exist symmetry outside the Euclidian group). First, let us define the patch image $\mathbf{x}$ in our dataset of patches $\mathcal{X}$ as $\mathbf{x} \in \mathcal{X}$. A patch pattern has symmetrical transformation $T$, if the patch is equal to itself when transformed or, $\mathbf{x} = T \mathbf{x}$. Common symmetrical transformations set $S$, which we desire, is a set of symmetrical transformations that are valid and true for every patch across the dataset. To be a symmetric pattern generation task, one common symmetrical transformation must atleast exist, or $S \neq \varnothing$. Finally, let $T_s$ be a symmetrical transformation within $S$ so the following properties hold true:


\begin{equation}
	\label{eq:one}
	\forall{{T}_{s} \in S}, \forall{\mathbf{x} \in \mathcal{X}}, (\mathbf{x} = T_{s} \mathbf{x})
\end{equation}

\subsubsection{Symmetric Pattern Enforcement}
\label{section:spe}
\begin{figure}
	\centering
	\includegraphics[width=\columnwidth]{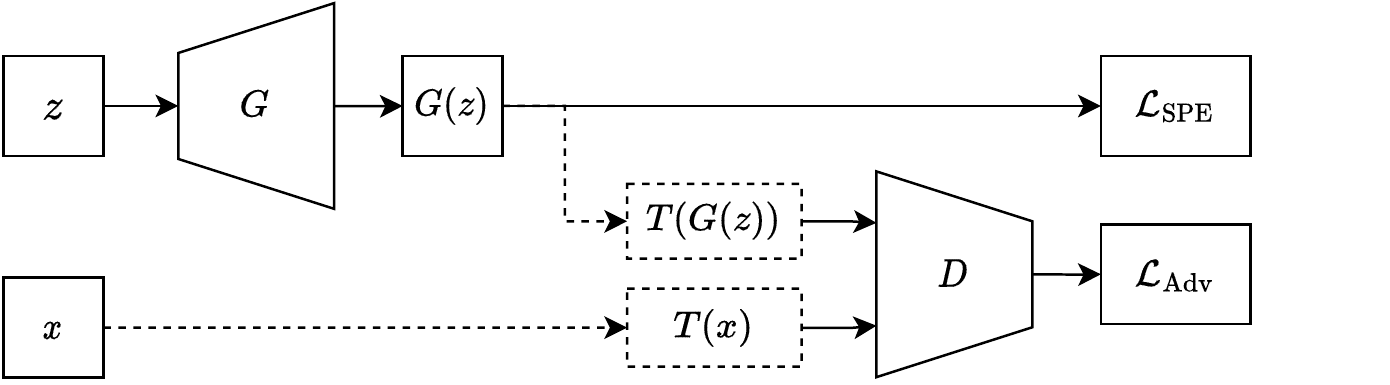}

	\caption[Symmetrical Pattern Enforcement loss overview]{Symmetrical Pattern Enforcement Enforcement loss overview. Let $z, G, G(z)$ be the latent input, generator, and generated image, with $x, D,$ and dashed line $ T$ as the real image, discriminator, and DiffAug operation. For $\mathcal{L}_{\text{SPE}}$ we directly take the generated image $G(z)$ before DiffAug applied. SPE loss unlike $\mathcal{L}_{\text{Adv}}$ adversarial loss only applied to the generator.}
	\label{fig:spe}
\end{figure}

Since we know every output image must have symmetric properties, by Eq. \ref{eq:one}, this motivates the idea to enforce symmetric properties for each pattern directly to the generator right from the beginning. We call this loss, \textbf{Symmetrical Pattern Enforcement (SPE)}, denotes by $\mathcal{L}_{\text{SPE}}$, where we aggregate image output similarity from the generator directly with what should have been its equivalent geometrical symmetry counterpart. Fig. \ref{fig:spe} shows where $\mathcal{L}_{\text{SPE}}$ used only the raw output from the generator and before \textit{DiffAug} $T$ augment both real and fake images. Lastly, to aggregate generated image with its symmetrical counterparts, our SPE Loss based on Similarity Loss $\mathcal{L}_{\text{Sim}}$. Thus, in general Symmterical Pattern Enforcement Loss formally defined as:

\begin{equation}
	\label{eq:two}
	\mathcal{L}_{\text{SPE}}(G(z)) = \frac{1}{|S|} \sum_{T_s \in S} \mathcal{L}_{\text{Sim}}(G(z), T_sG(z))
\end{equation}
where $G(z)$ is the generated output image directly taken from the generator, $\frac{1}{|S|}$ a normalization method for case of high cardinality of $S$.

\subsubsection{Implementation of Symmetric Pattern Enforcement for ITB-mBatik}
\label{section:spi}
\begin{figure}
	\centering
	\includegraphics[width=\columnwidth]{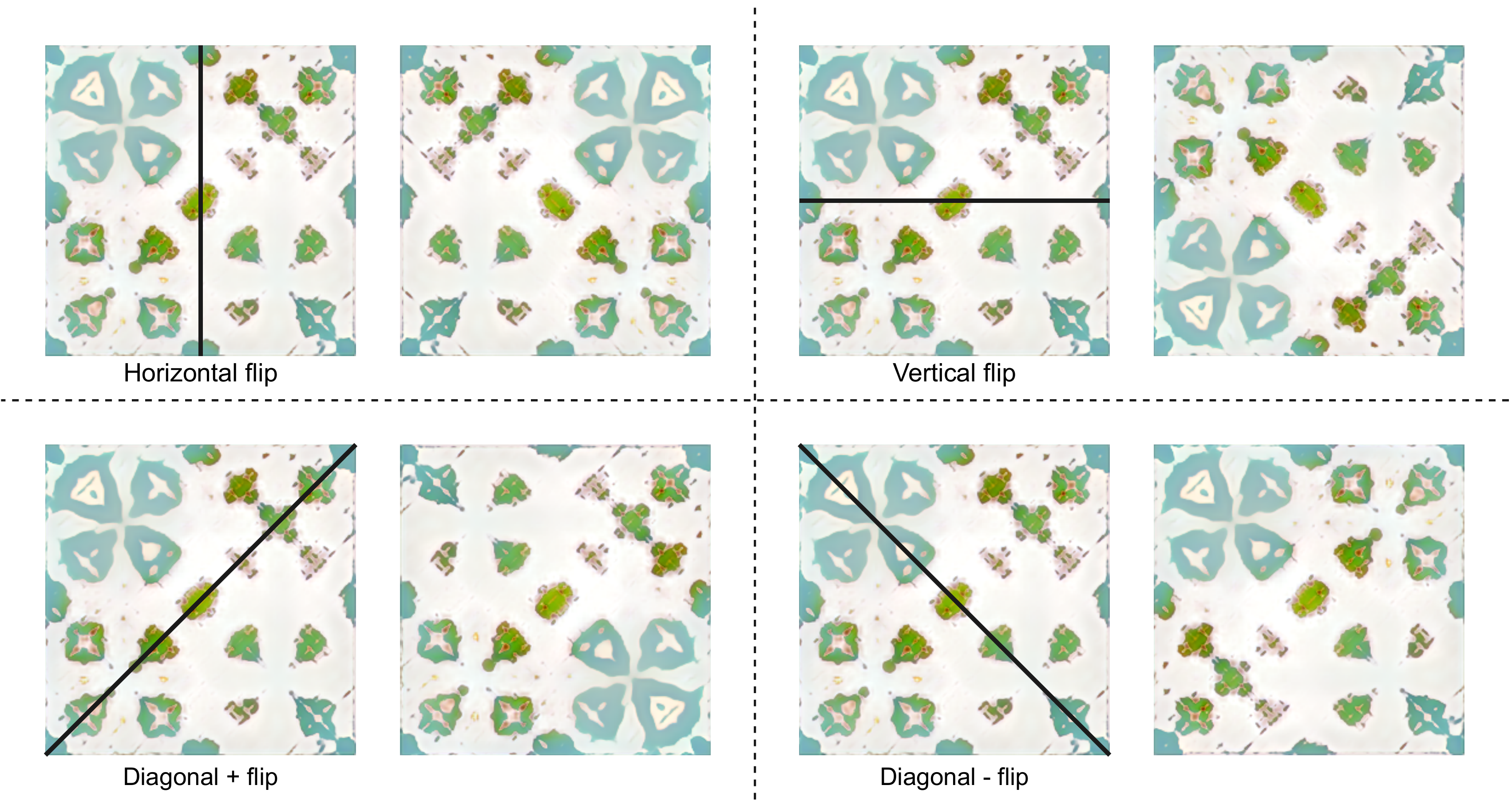}

	\caption[Symmetrical Pattern Enforcement operations ilustration]{Symmetrical Pattern Enforcement operations ilustration. All four operations we use on ITB-mBatik Dataset. Left image show the original image output from the generator with flip axis line and right image shows flip results where we compare both image similarity.}

	\label{fig:speillust}
\end{figure}
In our implementation, however, we reduce our generality for the sake of practicality. First, Cartan-Dieudonné theorem \cite{Gallier2001} establish that complex high dimensional isommetry can be constructed as a composition of reflections or multiple basic simple reflection. Therefor, in order to reduce loss complexity we will derive loss focusing on a set of observable intuitive reflection symmetry in ITB-mBatik Dataset.


We found 4 simple and most intuitive symmetrical transformation that satisfied the Eq. \ref{eq:one}, which are reflection on horizontal, vertical axis, and diagonal with positive gradient and negative counterpart as shown in Fig. \ref{fig:speillust}. Additionally, in application, the transformation operation from $R$ must be differentiable and support autograd for backprop. Therefore, let $R = \{\text{h-flip, v-flip, p-flip, n-flip}\}$, then we further optimize by finding the best subset operation by constructing powerset of $R$ or $\mathcal{P}(R)$. To reduce computation however, later we experiment with four subset for SPE Loss.

Next, using Eq. \ref{eq:two} as template, we implement SPE as:
\begin{align}
\mathcal{L}_{\text{SPE}}(G(z)) = \frac{1}{|R|} \sum_{T_r \in R} \mathcal{L}_{\text{Sim}}(G(z), T_rG(z))
\end{align}
Where for $\mathcal{L}_{\text{Sim}}$ we use simple similarities metrics L2, but $\mathcal{L}_{\text{Sim}}$ can use other function such as L1 or LPIPS\cite{zhang2018perceptual} as well.

\section{EXPERIMENTS}

\textbf{Dataset}. We used our ITB-mBatik dataset \cite{https://doi.org/10.17632/7hzr5539ws.1}. From the raw design files, we preprocess crop scale the repeating area ``patches'' to $1024 \times 1024$ pixels. Additionally, we handle translational symmetry by multi-phase sampling and increasing the Dataset training size.\\
\textbf{Metric}. In the evaluation process, we will measure five unsupervised image generation metrics. Fréchet Inception Distance \cite{NIPS2017_8a1d6947}, Improved Precision and Recall \cite{Kynkaanniemi2019}, and Density and Coverage \cite{ferjad2020icml} for better insight on generator performance.\\
\textbf{GAN Training Summary}. We used FastGAN with DiffAug as our baseline. We set output image size to $254 \times 254$ and for GANs losses, we believe different losses perform similarly from each other \cite{lucic2018gans}, therefore we use Hinge Loss \cite{lim2017geometric} for its simplicity and fast computation. For optimizers we use Adam Optimizers \cite{kingma2017adam} with learning rate of $2e-4$, $\beta_{1}=0.5, \beta_{2}=0.999$. We then train the GAN five times and report the best FID result. Each GAN trained for 200k minibatch time, where the batch size is 8. Lastly, for each 1k minibatch, we save the model for evaluation and possible early stopping purposes.\\
\textbf{Environment used}. For hardware, the research environment run on an i5-3470 CPU, 16GB of RAM, and GPU RTX 3060 with 12GB of VRAM. For software, the research environment will use PyTorch for its GAN model, which consists of the Generator and Discriminator, and TensorFlow to run InceptionV3 and VGG-16 for evaluation purposes.

\subsection{Symmetric Pattern Enforcement Analysis}

First, we experiment with $R:=$ \{hv, np, hvnp, [hv,np]\}, where h, v, n, p characters represent horizontal, vertical, diagonal negative, diagonal positive flip respectively, and [hv, np] is an operation where hv-flip, np-flip used interchangeably.

Table \ref{table:ressns} show the best result for each $R$ operation combination set. We find hv-flip combination achieves the lowest FID from 110.106 to 97.852 and increase baseline Recall by twice from 0.047 to 0.094, but there is slight deterioration on both Fidelity metric. Interestingly, however, if we use more symmetrical operation such as $hvnp$, SPE improve over baseline in every metric. In this case, however, since FID is the main metric to improve, further experimentation was done with SPE Loss config set to hv-flip.

\begin{table}
	\centering
	\small
	\caption[SPE Loss Experiment Results]{SPE Loss Experiment Results. \textbf{Bold} number show best score while \underline{underline} shows second bests.}
	\label{table:ressns}
	\begin{tabular}{llrrrrr}
		\toprule
		\multicolumn{2}{l}{\multirow{2}{*}{SPE Loss Config}} &       & \multicolumn{2}{c}{Fidelity} & \multicolumn{2}{c}{Diversity}\\
		\cmidrule(lr){4-5}\cmidrule(lr){6-7}
		\multicolumn{2}{l}{} & \multicolumn{1}{l}{FID$\downarrow$} & \multicolumn{1}{l}{Pre$\uparrow$} &  \multicolumn{1}{l}{Den$\uparrow$} & \multicolumn{1}{l}{Rec$\uparrow$} & \multicolumn{1}{l}{Cov$\uparrow$} \\
		\midrule

		\multicolumn{2}{l}{Baseline}        & 110.106         & 0.821                 &  0.863               & 0.047  & 0.557 \\  
		\midrule
		
		\multicolumn{1}{l}{\multirow{4}{*}{L2}}              & $\boldsymbol{hv}$          & \textbf{97.852} & 0.762 & 0.696 & \textbf{0.094} & 0.490\\
		\multicolumn{1}{c}{}                                 & $np$          & 103.782&\textbf{0.838}&\underline{0.866}&\underline{0.068}&\underline{0.564}\\
		\multicolumn{1}{c}{}                                 & $hvnp$        & \underline{102.624}&\underline{0.835}&\textbf{0.904}&0.067&\textbf{0.570}\\
		\multicolumn{1}{c}{}                                 & $[hv,np]$ & 111.291&0.819&0.762&0.048&0.463\\

		\bottomrule
	\end{tabular}

\end{table}



\subsection{Accelerating Attention Based Generator with SPE}
\label{section:accelerating}
\begin{table}
	\centering

	\caption[Attention based generator with SPE experiment results]{Attention based generator with SPE Enforcement experiment results. \textbf{Bold} number show best score while \underline{underline} shows second best. Second column next to attention mechanism name denotes on which feature resolution attention is applied to.}
	\label{table:res_finalattn}
	\begin{tabular}{llrrrrr}
		\toprule
		 \multicolumn{2}{l}{\multirow{2}{*}{Model}} & & \multicolumn{2}{c}{Fidelity} & \multicolumn{2}{c}{Diversity}\\
		\cmidrule(lr){4-5}\cmidrule(lr){6-7}
		           &     & \multicolumn{1}{l}{FID$\downarrow$} & \multicolumn{1}{l}{Pre$\uparrow$} &  \multicolumn{1}{l}{Den$\uparrow$} & \multicolumn{1}{l}{Rec$\uparrow$} & \multicolumn{1}{l}{Cov$\uparrow$} \\
		\midrule
		
		\multicolumn{2}{l|}{Baseline+SPE} & \underline{97.852} & 0.762 & 0.696 & 0.094 & \underline{0.490}\\
		\midrule

		\multicolumn{1}{l|}{\multirow{2}{*}{+SAN}} & \multicolumn{1}{l|}{$8^2$}  &122.27&\underline{0.797}&\underline{0.774}&0.011&0.428 \\
		\multicolumn{1}{l|}{}& \multicolumn{1}{l|}{$16^2$}& 139.78&\textbf{0.875}&0.773&0.007&0.298 \\
		\midrule

		\multicolumn{1}{l|}{\multirow{3}{*}{+EAttn}}& \multicolumn{1}{l|}{$8^2$} & 99.62&0.772&\underline{0.775}&0.087&0.461 \\
		\multicolumn{1}{l|}{}& \multicolumn{1}{l|}{$16^2$}      						 & 99.62&0.772&0.763&\underline{0.102}&0.480 \\
		\multicolumn{1}{l|}{}& \multicolumn{1}{l|}{$32^2$}      						 & \textbf{90.76}&0.737&0.661&\textbf{0.204}&\textbf{0.541} \\
		\bottomrule    
	\end{tabular}

\end{table}



\begin{figure}[t]
	\centering
	\includegraphics[width=\columnwidth]{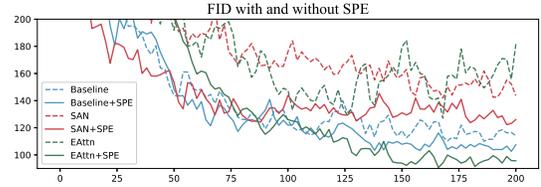}

	\caption[Attention based generator with SPE]{Attention based generator with SPE FID results. Each model trained for 200k iters. For clarity, we omit other worse performing attention on different resolution and put only the best performing SAN on $8^2$ and Efficient Attention on $32^2$ (more detail see Table \ref{table:res_finalattn}). }
	\label{fig:difreslapse}
\end{figure}



\begin{figure}[t]
	\centering
	\includegraphics[width=\columnwidth]{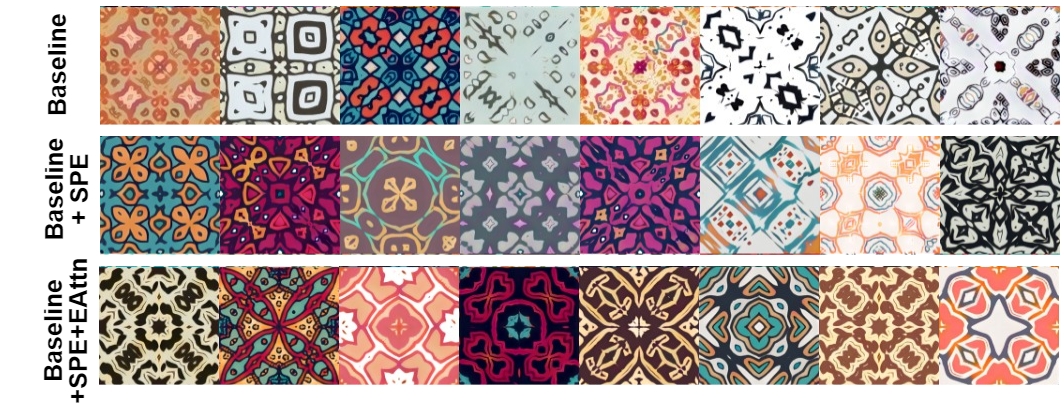}

	\caption[Generator Results difference between baseline and proposed method]{Generator Results difference between baseline and proposed method SPE with Efficient Attetion. Generator results are taken from each best FID snapshot, which is 186k for (a) Baseline, 200k for (b) Baseline+SPE, and 166k for (c) Baseline+SPE+EAttn.}
	\label{fig:resfinal}
\end{figure}

Attention mechanism achieves SOTA in many fields, such as attention-based GAN\cite{yu2021dual}, where attention mechanism Self Attention Network (SAN) \cite{Zhao_2020_CVPR} added into StyleGAN2 \cite{DBLP:journals/corr/abs-1912-04958}. However, the original attention based Generator was expensive computationally. Due to this, we also experiment with Efficient Attention (EAttn) \cite{shen2021efficient}, a more reasonable alternative for our limited settings.

In previous experimentation, attention-based generators have trouble training and converging due to increase in parameters and low training data size. However with SPE Loss, Fig. \ref{fig:difreslapse} shows every single model's performance improved significantly. Now, all EAttn generators from $8^2, 16^2, 32^2$ achieve lower than 100 FID with the smallest EAttn $32^2$ improved from 127.06 to a record low 90.76.

This record low FID however came with a tradeoff between fidelity and diversity. Every generator configuration that uses SPE, has a slightly reduced fidelity metrics. This however to be expected since FastGAN designed for high fidelity to begin with. In return when diversity increases we can see in Fig. \ref{fig:resfinal} between Baseline and Baseline with SPE+EAttn, noticeably SPE+EAttn have much finer details and better symmetry. We also find EAttn covers more different styles with better detail than the previous method.



\section{CONCLUSION}
With the massive stride of advancements in GAN, particularly in limited computation and data, mainstream applications of GAN will start to follow. As such, this paper covers the batik generation task, a pop icon in Southeast Asia. While doing this research, we found symmetrical pattern tasks generalizable to similar tasks, such as Portuguese or Morocan tiles. With a new and better high-quality dataset, exploration of cheaper attention adoption to GAN in limited computation, and SPE Loss for training acceleration, we improve symmetric pattern generation significantly from the best state-of-the-art FastGAN in terms of image quality (FID) and detail (recall). We believe this research would be the start and the foundation of future AI Art, Symmetric Pattern Generation applications for artists.


\bibliographystyle{IEEEbib}
\bibliography{references}

\end{document}